\documentclass[11pt]{article}
\usepackage[margin=1in]{geometry}
\usepackage[T1]{fontenc}
\usepackage[utf8]{inputenc}
\usepackage{lmodern}
\usepackage{microtype}
\usepackage{amsmath,amssymb}
\usepackage{graphicx}
\usepackage{booktabs}
\usepackage{caption}
\usepackage[american]{babel}
\usepackage{csquotes}
\usepackage{url}
\usepackage[authoryear,round]{natbib}
\usepackage[hidelinks]{hyperref}
\usepackage{orcidlink}
\usepackage{setspace}
\newcommand{\pkg}[1]{\textbf{#1}}
\newcommand{\code}[1]{\texttt{#1}}

\title{Model or Scaffold? Decomposing the Ability of AI Coding Agents\\ with Explanatory Item Response Theory}
\author{Okan Bulut\,\orcidlink{0000-0001-5853-1267}\thanks{Correspondence concerning this article should be
addressed to Okan Bulut, Centre for Research in Applied Measurement and Evaluation, University of Alberta,
6-110 Education Centre North, 11210 87 Ave NW, Edmonton, AB T6G 2G5, Canada. Email: \href{mailto:bulut@ualberta.ca}{bulut@ualberta.ca}}\\[0.4em]
\normalsize Centre for Research in Applied Measurement and Evaluation, University of Alberta}
\date{November 2026}

\begin{document}
\maketitle

\begin{abstract}
\noindent
A coding agent is a language model inside a scaffold, the program that gives the model its tools,
prompts and retries. Leaderboards such as SWE-bench Verified report one resolve rate per agent and
cannot say how much of it belongs to the model and how much to the scaffold. Recent work fits item
response models in which an agent's ability is the sum of a model ability and a scaffold ability.
We frame the question as a cross-classified explanatory item response model
\citep{deboeck2004explanatory, vandennoortgate2003cross}: Model, scaffold and the particular
model--scaffold pair are crossed random effects on ability, tasks are random items, and task features
explain difficulty. Fitted as a generalized linear mixed model with standard software, the model
gives a variance partition, a likelihood ratio test of additivity, and standard errors for every
ability and every prediction. We apply it to 166 public runs of SWE-bench Verified (83,000 task
results), to 143 runs of Terminal-Bench 2.0 with repeated attempts (62,682 attempts), and to
simulated leaderboards. On all SWE-bench Verified runs, 53\% of the ability variance sits with the
model, 38\% with the scaffold and 9\% with the pair, but 93 of 111 models appear in only one scaffold,
so model and scaffold are largely confounded. Where models and scaffolds cross, the model carries 81\%,
the scaffold 14\% and the pair 5\%. Terminal-Bench 2.0, whose leaderboard runs every submission in one
harness with five attempts per task, gives the same picture: 88\% model, 8\% scaffold and 4\% pair. On
both benchmarks the pair term is statistically needed, so ability is not strictly additive, but the
departure is small. A model's effect estimated without a fixed reference scaffold correlates .95 with
its ability in that scaffold, and a held-out run is predicted from its model and scaffold within 4 to 7
percentage points. Adaptive testing with the fitted difficulties recovers full-benchmark ability from a
few dozen tasks (25 tasks: $r = .98$). Everything runs in the R package \pkg{mirtverse}.

\end{abstract}

\section{Introduction}

Software-engineering benchmarks have become the main way to compare AI coding agents. In SWE-bench
\citep{jimenez2024swebench}, each task is a real issue from an open-source Python project, and an
agent resolves the task when its patch makes the project's hidden tests pass. The human-validated
subset SWE-bench Verified \citep{openai2024verified} has 500 tasks, and its public leaderboard
lists one resolve rate per submitted run. The SWE-bench team also publishes the task-by-task results
of every run \citep{swebench2026experiments}, which makes the leaderboard a response matrix: Runs
are test takers, tasks are items, and each cell is a 0 or 1.

A run is not a single system. It pairs a \emph{language model}, which reads the issue and writes code,
with a \emph{scaffold}, which decides what the model sees, which tools it may call, how often it may
retry and how a patch is chosen. SWE-agent \citep{yang2024sweagent}, OpenHands
\citep{wang2025openhands} and dozens of in-house systems are scaffolds. The same model can score
very differently in two scaffolds, and leaderboards that rank runs cannot tell a reader whether a
higher score reflects a better model or a better scaffold \citep{kapoor2024matter,
kapoor2026hal}. That question matters for anyone choosing a model, building a scaffold or
interpreting progress.

Item response theory (IRT) has a long history of separating people from items
\citep{rasch1960, lord1980}, and it is now widely used in AI evaluation to weight items by
difficulty, shorten benchmarks and test models adaptively \citep{lalor2016building,
martinezplumed2019irt, rodriguez2021evaluation, polo2024tiny, hofmann2025fluid, zhou2026lost}.
Two recent papers bring the model--scaffold question into this frame. \citet{ge2026agent} write an
agent's ability as the sum of a model ability and a scaffold ability, fit the model by variational
inference across four coding benchmarks, predict held-out model--scaffold combinations, and use the
prediction as a prior for adaptive testing. They report that the two parts combine additively without
pairwise interactions. \citet{liu2026converged} audit the same SWE-bench submissions with
paired tests on cell means and find that scaffold orderings reverse across models, with six of nine
interaction contrasts significant.

These findings point in different directions, and both rest on a design that is mostly not crossed:
Most models on the leaderboard were run in a single scaffold, and most scaffolds were run with very
few models. This paper treats the question as a standard measurement problem. The additive
decomposition of \citet{ge2026agent} is a special case of a cross-classified explanatory item
response model \citep{vandennoortgate2003cross, deboeck2004explanatory}, which psychometrics has
used for two decades to study persons nested in several groupings at once. Written that way, the
model can be fitted as a generalized linear mixed model with \pkg{lme4} \citep{bates2015lme4}
through the \pkg{eirm} package \citep{bulut2021eirm}, and it delivers four things the earlier
work does not report together:

\begin{enumerate}
\item A \emph{variance partition} of agent ability into model, scaffold and pair components, with the
pair component playing the role of the model $\times$ scaffold interaction.
\item A \emph{likelihood ratio test of additivity}: Dropping the pair term gives the purely additive
model, so additivity is a testable restriction rather than an assumption.
\item \emph{Identification diagnostics}: We locate the connected core of the leaderboard where model and
scaffold actually cross, and compare the partition there with the partition on all runs.
\item \emph{Uncertainty}: Every ability, every model and scaffold effect and every prediction for an
unseen pair comes with a standard error, and we check the coverage of the resulting intervals on
held-out runs.
\end{enumerate}

We report a simulation study of parameter recovery (Study~1), an analysis of 166 public runs of
SWE-bench Verified (Study~2), and a replication on Terminal-Bench 2.0 (Study~3), a benchmark whose
leaderboard uses one harness, repeated attempts and a fixed reference scaffold. The functions used here, \code{measure\_agents()},
\code{predict()}, \code{agent\_cat()} and \code{agent\_cat\_check()}, are part of the R package
\pkg{mirtverse} \citep{bulut2026mirtverse}.

\section{Related Work}

\paragraph{IRT for AI Evaluation.} \citet{lalor2016building} first used IRT abilities in place of
accuracy for NLP models, and \citet{martinezplumed2019irt} applied IRT to classifiers at the instance
level. \citet{rodriguez2021evaluation} showed that difficulty and discrimination make leaderboards
more reliable. \citet{polo2024tiny} used IRT to estimate full-benchmark scores from about a hundred
items, and \citet{hofmann2025fluid} combined IRT with adaptive item selection, which lowered variance
and raised validity. \citet{zhou2026lost} examined benchmark quality with richer IRT models.
\citet{jiang2026trust} warn that AI benchmarks invert the usual psychometric regime, with few test
takers and many items, and that item parameters are poorly recovered with 30 test takers; they
recommend checking recovery by simulating from one's own fitted model, which we do in Study~1.

\paragraph{Agents and Scaffolds.} \citet{ge2026agent} introduced the additive model--scaffold
decomposition summarized above, together with task features drawn from issue text, patches and tests.
\citet{ndzomga2026efficient} studied task subsets for agent benchmarks under scaffold shift and
found that rankings survive where absolute scores do not. \citet{liu2026converged} showed that the
top of SWE-bench Verified can no longer be ordered by paired tests, reconstructed a model $\times$ scaffold
design from the run metadata, and found within-model scaffold ranges of up to 30 percentage points.
Our work differs from these in using a single likelihood-based model that treats model, scaffold and
pair as random effects, which gives a variance partition, a formal test of additivity and standard
errors in one fit.

\section{The Model}

Let $y_{aj} \in \{0, 1\}$ record whether run $a$ resolved task $j$. Each run uses model $m(a)$ and
scaffold $s(a)$. The cross-classified explanatory model is
\begin{equation}
\operatorname{logit} P(y_{aj} = 1) = \beta_0 + \underbrace{\mu_{m(a)} + \gamma_{s(a)} + \delta_a}_{\theta_a}
 - \mathbf{x}_j^\top \boldsymbol{\beta} - \varepsilon_j,
\label{eq:model}
\end{equation}
with $\mu_m \sim N(0, \sigma^2_\mu)$, $\gamma_s \sim N(0, \sigma^2_\gamma)$, $\delta_a \sim N(0,
\sigma^2_\delta)$ and $\varepsilon_j \sim N(0, \sigma^2_\varepsilon)$, all independent. The ability of
run $a$ is $\theta_a = \mu_{m(a)} + \gamma_{s(a)} + \delta_a$. The task's difficulty is
$b_j = -\beta_0 + \mathbf{x}_j^\top\boldsymbol{\beta} + \varepsilon_j$, in which task features
$\mathbf{x}_j$ (here the repository) explain part of the difficulty and $\varepsilon_j$ is the rest
\citep{deboeck2004explanatory}. Without $\mu$, $\gamma$ and features, Equation~\ref{eq:model} is
the Rasch model with random persons and random items. When a run attempts a task $n_{aj}$ times, as in
Study~3, $y_{aj}$ counts the successes and the same model is binomial.

\paragraph{The Pair Term.} $\delta_a$ is the part of a run's ability that its model and scaffold do not
explain. In a design where each pair is run once, it is the model $\times$ scaffold interaction plus
run-to-run noise, plus anything else that differs between runs with the same labels, such as a newer
scaffold version or a different number of attempts. The purely additive model of
\citet{ge2026agent} sets $\sigma^2_\delta = 0$; this is a restriction on a variance parameter, and a
likelihood ratio test of it is conservative because the null value lies on the boundary
\citep{self1987boundary}.

\paragraph{The Variance Partition.} The shares $\sigma^2_\mu$, $\sigma^2_\gamma$ and $\sigma^2_\delta$
of $\sigma^2_\mu + \sigma^2_\gamma + \sigma^2_\delta$ say how much of the spread in ability comes from the
model, the scaffold and the pair. They are only meaningful where the design separates the three: If a
scaffold is run with one model only, its $\gamma_s$ and that model's $\mu_m$ are identified only through
the shrinkage of the random-effects prior. We therefore also fit the model on the \emph{connected core},
obtained by repeatedly removing runs whose model appears in only one scaffold or whose scaffold appears
with only one model, until nothing changes.

\paragraph{Estimation.} We fit Equation~\ref{eq:model} by Laplace-approximated maximum likelihood with
\code{lme4::glmer} \citep{bates2015lme4} called through \code{eirm::eirm} \citep{bulut2021eirm},
using the \code{nloptwrap} optimizer, which gave the same estimates as the default optimizer several times
faster. The model, scaffold and pair effects are the conditional modes. Because these do not carry the
uncertainty in the task difficulties, the reported standard error of a run's ability is the posterior
standard deviation of $\theta_a$ given its own results, the fitted difficulties and a normal prior with
the mean and SD of the fitted abilities (an expected a posteriori estimate on a grid).

\paragraph{Predicting an Unseen Pair.} For a model and scaffold that were run separately but never
together, the predicted ability is $\hat\mu_m + \hat\gamma_s$, and its standard error adds the conditional
variances of the two effects and $\hat\sigma^2_\delta$, since the new pair's own $\delta$ is unknown. A
model or scaffold never seen gets a zero effect and its full variance. The predicted resolve rate averages
$\operatorname{logit}^{-1}(\hat\theta - \hat b_j)$ over tasks.

\paragraph{Adaptive Testing.} A new run can be tested on a few tasks chosen one at a time
\citep{weiss1984cat, vanderlinden2010cat}. After each result we update the expected a posteriori
ability, and the next task is the unused task whose difficulty is closest to the current estimate,
which maximizes Fisher information under the Rasch model. \code{agent\_cat\_check()} replays this for
every measured run with its own results and compares the short adaptive test with random task sets of
the same length.

\section{Data}

We read every run in the \code{evaluation/verified} folder of the SWE-bench experiments repository
\citep{swebench2026experiments} in November 2026: 168 runs submitted between October 2023 and
September 2026, each scored on all 500 tasks. A task that a run did not attempt or did not produce a patch
for counts as unresolved, the convention the leaderboard uses. As a quality check, we compared each run's
resolve rate computed from its task-level results with the resolve rate reported in its metadata, which 41 runs
give. All of these agreed within one percentage point except two whose task-level files are broken: A Gemini 3 Pro run
in mini-swe-agent reports 69.6\% but marks every task unresolved, and a Claude 3.7 Sonnet run in
mini-swe-agent reports 52.8\% but marks 89.8\% of tasks unresolved. We dropped both, which leaves 166 runs
and 83,000 results. The mean resolve rate is .531, runs range from .004 to .792, and 27 tasks were
resolved by no run.

Each run's \code{metadata.yaml} names its model and its scaffold, but not in a standard form
\citep[see also][]{liu2026converged}. We lowercased names, removed organization prefixes, URLs and
date suffixes, wrote version numbers with hyphens, mapped spellings such as \code{claude-4-sonnet} to
\code{claude-sonnet-4}, and stripped version numbers from scaffold names, so that a scaffold here is a
scaffold family. Runs that do not disclose the model (34 runs) get a model of their own, and runs that
use several models (23 runs) are treated as one ensemble model. The cleaned data have 111 models (77
disclosed single models) and 65 scaffolds. The design is sparse: 93 of the 111 models appear in a single
scaffold, 39 of the 65 scaffolds were run with a single model, and 34 scaffolds have a single run.
mini-swe-agent, a fixed reference scaffold, accounts for 39 runs. Twelve model--scaffold cells hold
more than one run. The connected core has 35 runs, 10 models and 10 scaffolds.

\section{Study 1: Recovery in Simulated Leaderboards}

\paragraph{Design.} Following \citet{jiang2026trust} and \citet{morris2019sim}, we simulated
leaderboards close to the real one and checked what the model recovers. Each leaderboard had 15 models
and 6 scaffolds ($\sigma_\mu = 1$), task difficulties with SD 1.5 that depend on a repository and a size
feature, and four crossed factors: 200 or 500 tasks; 30\% or 60\% of the 90 model--scaffold pairs run
(each model and each scaffold run at least once); scaffold SD 0.3 or 0.8; and pair SD 0.3 or 1.0. Each of
the 16 conditions was replicated 10 times. We report the correlation of the estimated model and scaffold
effects and task difficulties with the truth, the correlation and coverage of predicted abilities for the
pairs that were not run, and the error in their predicted resolve rate.

\begin{figure}[t]
\centering
\includegraphics[width=\linewidth]{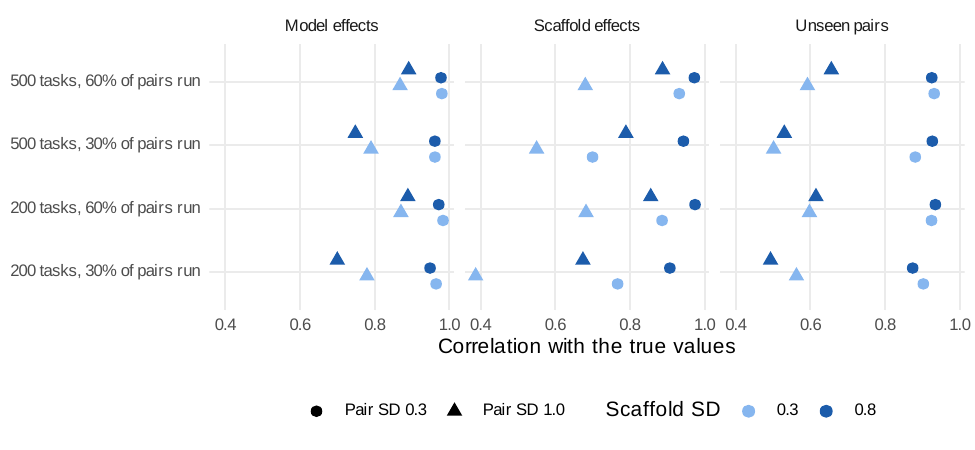}
\caption{Simulation: Correlation of estimated model effects, scaffold effects and predicted abilities of pairs that were not run with their true values, by condition.}
\label{fig:sim}
\end{figure}

\paragraph{Results.} All 160 fits converged, in 7 seconds on average. Table~\ref{tab:sim} summarizes recovery by the two variance factors, which mattered most. Abilities of the runs that were made (correlation .99), task difficulties (.96) and the size effect on difficulty (mean estimate 0.60, true 0.6) were recovered well everywhere. The variance components were close to their true values: For example, with true scaffold and pair variances of 0.64 and 1.00, the mean estimates were 0.75 and 0.83. Model and scaffold effects were recovered well when the pair SD was small (.97 and .82 to .95) but less well when it was large (.82 to .84 and .57 to .81), and scaffold effects were the hardest to recover when scaffolds differed little. The pair SD set the ceiling for predicting pairs that were not run: With a pair SD of 0.3, predicted abilities correlated .91 with the truth and the predicted resolve rate was off by about 5 points; with a pair SD of 1.0, the correlation fell to .58 and the error rose to 14 to 15 points. Running a larger share of pairs helped (Figure~\ref{fig:sim}), and more tasks helped little. The 95\% intervals for unseen pairs covered the truth 82\% to 92\% of the time, somewhat below the nominal level, because the conditional variances of the effects understate their uncertainty when there are only 6 scaffolds.

\begin{table}[t]
\centering
\caption{Simulation recovery, averaged over number of tasks and share of pairs run (40 replications per row). True model variance is 1.}
\label{tab:sim}
\small
\begin{tabular}{rrrrrrrrrr}
\toprule
\multicolumn{2}{c}{True SD} & \multicolumn{3}{c}{Estimated variance} & \multicolumn{2}{c}{Correlation with truth} & \multicolumn{3}{c}{Unseen pairs} \\
\cmidrule(lr){1-2}\cmidrule(lr){3-5}\cmidrule(lr){6-7}\cmidrule(lr){8-10}
Scaffold & Pair & Model & Scaffold & Pair & Model & Scaffold & $r$ & MAE & Coverage \\
\midrule
0.3 & 0.3 & 0.98 & 0.08 & 0.09 & .97 & .82 & .91 & 5.2 & .88 \\
0.8 & 0.3 & 0.86 & 0.53 & 0.10 & .97 & .95 & .92 & 5.4 & .82 \\
0.3 & 1.0 & 1.01 & 0.14 & 0.92 & .84 & .57 & .58 & 13.8 & .92 \\
0.8 & 1.0 & 1.11 & 0.75 & 0.83 & .82 & .81 & .59 & 14.8 & .90 \\
\bottomrule
\end{tabular}
\end{table}

The simulation has two lessons for the real data. The model recovers what it should when the design is crossed, including small pair variances, so the variance partitions in Studies~2 and~3 are not estimation artifacts. And the size of the pair SD decides how well a new pair can be predicted: The real pair SDs (0.3 to 0.6 logits) are close to the favorable condition.

\section{Study 2: SWE-bench Verified}

\subsection{Does Knowing the Model and Scaffold Help?}

Table~\ref{tab:fits} compares three nested models. Adding model and scaffold effects to the Rasch model
improved fit ($\chi^2(2) = 116.0$, $p < .001$; AIC and BIC both lower), and adding the repository as a task
feature improved it further ($\chi^2(11) = 35.3$, $p < .001$), although BIC prefers the model without it.
The repository explains 7\% of the variance in task difficulty: Scikit-learn tasks are clearly easier than the
astropy baseline ($p < .001$), and the other repositories differ less clearly. Ability estimates are very
precise (median SE 0.14 logits, reliability .99), because each run took all 500 tasks. Each fit took a few
minutes.

\begin{table}[t]
\centering
\caption{Model comparison on 166 runs $\times$ 500 tasks.}
\label{tab:fits}
\begin{tabular}{lrrr}
\toprule
Model & Parameters & AIC & BIC \\
\midrule
M0: Rasch (run and task random) & 3 & 55,118.2 & 55,146.2 \\
M1: + model and scaffold & 5 & 55,006.2 & 55,052.9 \\
M2: + repository & 16 & 54,992.9 & 55,142.1 \\
\bottomrule
\end{tabular}
\end{table}

A 2PL model fitted with \pkg{mirt} \citep{chalmers2012mirt} to the 473 tasks that some but not all
runs resolved fits better than the Rasch model by AIC (50,663 against 53,181) and BIC (53,607 against
54,656), and the median discrimination is 2.3. Abilities from the two models correlate .995, and Rasch
abilities from \pkg{mirt} correlate .999 with the M1 abilities, so the Rasch form costs little for ranking
runs; we keep it because it is what the cross-classified model needs.

\subsection{How Is Ability Variance Shared?}

\begin{figure}[t]
\centering
\includegraphics[width=\linewidth]{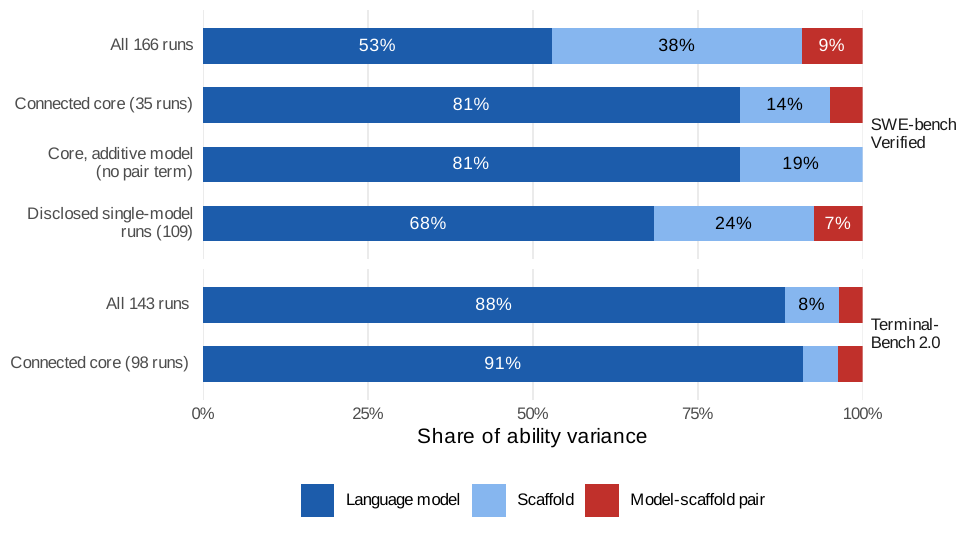}
\caption{Share of ability variance that belongs to the language model, the scaffold and the particular
model--scaffold pair. SWE-bench Verified: All runs, the connected core where models and scaffolds cross, the
core without a pair term, and runs with one disclosed model. Terminal-Bench 2.0 (Study~3): All runs and the
connected core.}
\label{fig:variance}
\end{figure}

Figure~\ref{fig:variance} shows the partition. On all 166 runs, model differences account for 53\% of
the ability variance ($\hat\sigma^2_\mu = 1.66$), scaffold differences for 38\% ($\hat\sigma^2_\gamma =
1.19$) and the pair for 9\% ($\hat\sigma^2_\delta = 0.29$). The scaffold share is hard to trust. With 93
models in one scaffold and 39 scaffolds with one model, most runs are the only evidence for both their model
and their scaffold, and the split between $\mu$ and $\gamma$ for those runs is set mostly by the
random-effects prior, not by the data.

In the connected core, where every model was run in at least two scaffolds and every scaffold with at
least two models, the model carries 81\% of the variance ($\hat\sigma^2_\mu = 5.11$), the scaffold 14\%
($\hat\sigma^2_\gamma = 0.85$) and the pair 5\% ($\hat\sigma^2_\delta = 0.31$). With only 10 models and 10 scaffolds these shares are imprecise: Parametric bootstrap intervals (100
replications) run from .54 to .93 for the model, .02 to .39 for the scaffold and .02 to .14 for the pair, and
a second optimizer gave the same estimates. Likelihood ratio tests
in the core (Table~\ref{tab:core}) agree: Dropping the model term costs $\chi^2(1) = 24.7$ and dropping the
pair term $\chi^2(1) = 184.9$ (both $p < .001$), while the scaffold term is not significant
($\chi^2(1) = 2.6$, $p = .11$; the test is conservative at the boundary). Without the pair term, the additive
model puts 19\% of the variance on scaffolds, so an additive fit moves some pair-specific variance onto
scaffolds. The partition is not an artifact of undisclosed or ensemble runs: Refitting M1 on the 109 runs with
one disclosed model (59 models, 39 scaffolds) gives 68\% model, 24\% scaffold and 7\% pair.

\begin{table}[t]
\centering
\caption{Likelihood ratio tests in the connected core (35 runs, 10 models, 10 scaffolds). Each row
drops one term from the full model with model, scaffold, pair and task effects.}
\label{tab:core}
\begin{tabular}{lrrr}
\toprule
Term dropped & $\chi^2$ & df & $p$ \\
\midrule
Scaffold & 2.6 & 1 & .11 \\
Model & 24.7 & 1 & $< .001$ \\
Model--scaffold pair & 184.9 & 1 & $< .001$ \\
\bottomrule
\end{tabular}
\end{table}

\begin{figure}[t]
\centering
\includegraphics[width=0.85\linewidth]{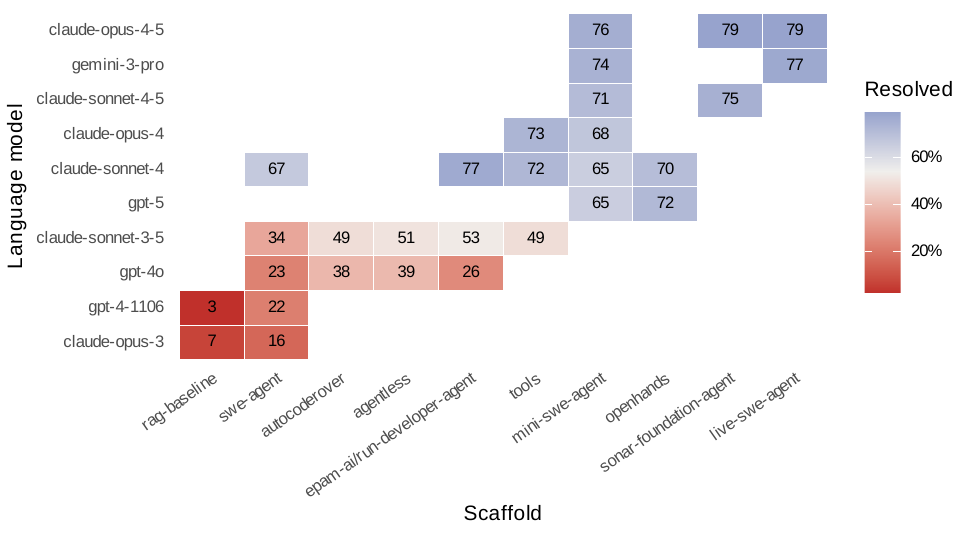}
\caption{Resolve rates (\%) in the connected core, averaged over runs in a cell. Rows and columns are
ordered by their mean. Empty cells were never run.}
\label{fig:core}
\end{figure}

On all 166 runs, the additive model is also rejected: Dropping the pair term from M1 raises the deviance by
228 on one degree of freedom. The pair SD is 0.54 logits, against 1.29 for models. Figure~\ref{fig:core} shows
both patterns. Rows differ much more than columns, so the model matters most, but the scaffold that helps one
model does not always help another: EPAM's agent gave Claude 3.5 Sonnet its best result (53\%) and GPT-4o one of
its worst (26\%, against 38\% to 39\% in Agentless and AutoCodeRover). Among model pairs run in the same two
scaffolds, the ordering of the two models reversed in 2 of 15 comparisons, both with small gaps.

\subsection{Predicting Runs Nobody Has Made}

We split the 166 runs into five folds at random, refitted M1 on four folds, and predicted each held-out run from its model and scaffold alone, before seeing any of its results. Held-out runs fall into four groups: Runs whose exact pair is also in the training folds (24 runs, mostly repeated submissions), runs whose model and scaffold were both seen but never together (36), runs with one of the two seen (83), and runs with neither seen (23). We compare the predicted resolve rate with three baselines: The grand mean of training runs, the mean of training runs with the same model, and an additive rule on rates (model mean plus scaffold mean minus grand mean).

\begin{figure}[t]
\centering
\includegraphics[width=\linewidth]{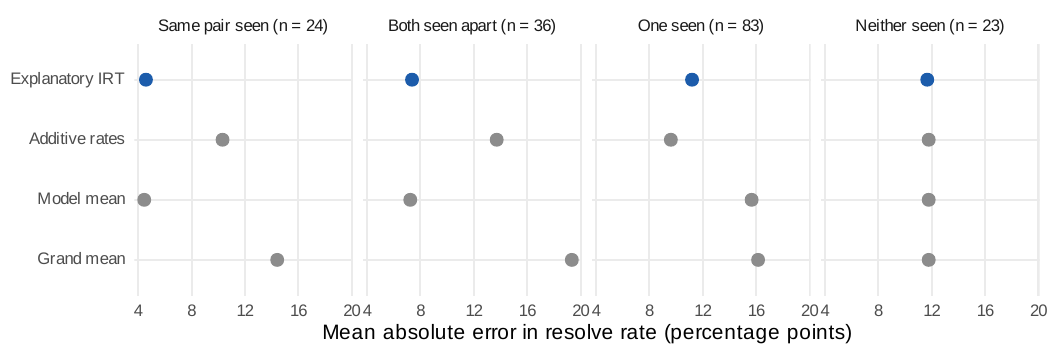}
\caption{Error in the predicted resolve rate of held-out runs, by what the training folds had seen.}
\label{fig:holdout}
\end{figure}

Table~\ref{tab:holdout} and Figure~\ref{fig:holdout} show the results. For the 36 unseen pairs whose model and scaffold were both seen, the explanatory model missed the resolve rate by 7.4 points on average, about as well as the mean of the same model's other runs (7.2) and much better than the additive rule on rates (13.7) or the grand mean (19.3). Its 95\% intervals for these runs' abilities covered the full-benchmark ability in 94\% of cases. When only one of the two was seen, the error rose to 11.2 points, and with neither seen the model can only predict the average. For repeated runs of a pair already in the training data, the prediction is close (4.6 points), but its interval, which uses the earlier run's small standard error, covered only 58\% of the time: Two submissions with the same labels differ more than measurement error allows, so part of $\delta_a$ is run-to-run and version-to-version variation.

Task-level log loss tells the same story. For unseen pairs, the model's predicted probabilities beat the task's training resolve rate (0.35 against 0.49 per task), because they combine the task's difficulty with the run's predicted level. Within a run, however, the model orders tasks only by difficulty, so its within-run AUC (.92) is the same as that of the task resolve rate; most of the discrimination in task-level prediction comes from the tasks, not from knowing the agent.

\begin{table}[t]
\centering
\caption{Held-out runs, five folds. MAE is the mean absolute error in resolve rate (percentage points); coverage is the share of 95\% intervals for ability that contain the full-benchmark ability.}
\label{tab:holdout}
\small
\begin{tabular}{lrrrrrrr}
\toprule
 & & \multicolumn{4}{c}{MAE in resolve rate} & \multicolumn{2}{c}{} \\
\cmidrule(lr){3-6}
Held-out run & $n$ & Grand & Model & Additive & IRT & Log loss & Coverage \\
\midrule
Same pair in training & 24 & 14.4 & 4.5 & 10.3 & 4.6 & 0.34 & .58 \\
Model and scaffold seen apart & 36 & 19.3 & 7.2 & 13.7 & 7.4 & 0.35 & .94 \\
One of them seen & 83 & 16.1 & 15.7 & 9.6 & 11.2 & 0.40 & .96 \\
Neither seen & 23 & 11.8 & 11.8 & 11.8 & 11.7 & 0.44 & 1.00 \\
\bottomrule
\end{tabular}
\end{table}

\subsection{Testing a New Run on a Few Tasks}

\begin{figure}[t]
\centering
\includegraphics[width=\linewidth]{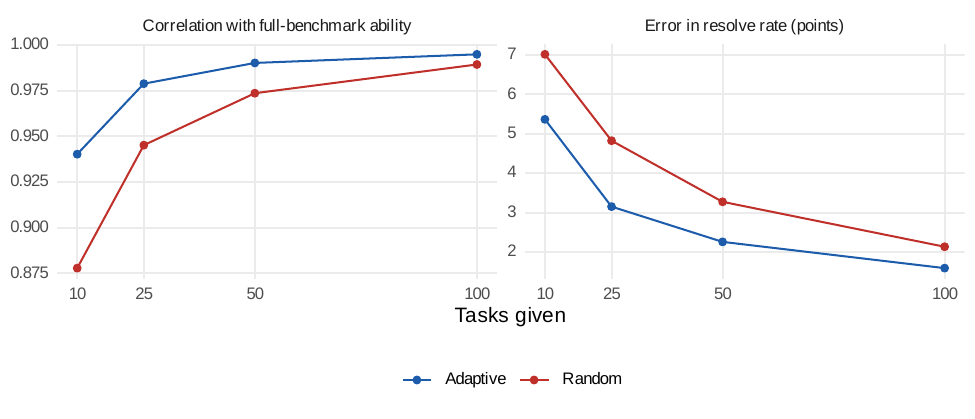}
\caption{Short tests of held-out runs. Left: Correlation of abilities from the short test with abilities from all 500 tasks. Right: Mean absolute error of the expected resolve rate, in percentage points. Random tests are scored with the same model and averaged over five draws.}
\label{fig:cat}
\end{figure}

For each held-out run we also ran an adaptive test on its own results, using the task difficulties from the other four folds, and compared it with random task sets of the same length scored by the same model (Figure~\ref{fig:cat}). Ten adaptive tasks gave abilities that correlated .94 with the full-benchmark abilities, against .88 for ten random tasks; 25 adaptive tasks reached .98 (random: .95), and 50 reached .99 (random: .97). The expected resolve rate from 25 adaptive tasks was within 3.2 points of the run's real resolve rate on average, against 4.8 points for 25 random tasks scored by the model and 7.0 points for the raw proportion resolved in those 25 tasks. A run that costs hundreds of dollars on all 500 tasks can thus be placed on the leaderboard's scale from a few dozen, provided the leaderboard's difficulties are available, as \citet{ge2026agent} and \citet{hofmann2025fluid} also found.

\section{Study 3: Terminal-Bench 2.0}

\paragraph{Data.} Terminal-Bench 2.0 \citep{merrill2026terminal} has 89 tasks done in a computer terminal,
from compiling software to cryptanalysis. Its leaderboard runs every submission through one evaluation harness,
asks for five attempts per task, and its authors ran many models themselves in a fixed reference scaffold,
Terminus 2. We used the per-attempt results that \citet{ge2026agent} collected from the leaderboard and
released under the MIT license: 143 runs submitted between October 2025 and May 2026, with 62,682 attempts
(47.2\% successful). Most runs made five attempts per task. We cleaned model and scaffold names as in Study~2
and gave each of the 9 runs with several models a model of its own. The design is much better crossed than
SWE-bench Verified: 29 of the 55 models appear in more than one scaffold, 23 of the 45 scaffolds were run with
more than one model, and Terminus 2 accounts for 33 runs. The connected core keeps 98 runs, 26 models and 18
scaffolds. We fitted Equation~\ref{eq:model} with binomial counts and no task features; each fit took a few
seconds.

\paragraph{Variance Partition.} Adding model and scaffold effects to the Rasch model improved fit greatly
($\chi^2(2) = 212.8$). On all runs, the model carries 88\% of the ability variance ($\hat\sigma^2_\mu = 2.40$),
the scaffold 8\% ($\hat\sigma^2_\gamma = 0.22$) and the pair 4\% ($\hat\sigma^2_\delta = 0.10$); in the
connected core the shares are 91\%, 5\% and 4\% (Figure~\ref{fig:variance}). All three terms are needed in
the core: Dropping the scaffold costs $\chi^2(1) = 15.3$, the model $\chi^2(1) = 145.4$ and the pair
$\chi^2(1) = 181.2$ (all $p < .001$). So additivity is rejected here too, but the pair SD is 0.31 logits, close
to the 0.56 of the SWE-bench core, and an additive model is a good approximation. Parametric bootstrap intervals (100 replications) for the shares on all runs were .82 to .93 for the
model, .04 to .14 for the scaffold and .02 to .06 for the pair. The partition barely moved with another
optimizer (\code{bobyqa}), with the uncleaned names (87\%, 10\%, 4\%), without ensemble runs (87\%, 8\%, 4\%),
with only the 117 runs that made five attempts on at least 90\% of tasks (89\%, 7\%, 4\%), or with majority-vote
binary scoring as in \citet{ge2026agent} (89\%, 8\%, 3\%).

\begin{figure}[t]
\centering
\includegraphics[width=0.6\linewidth]{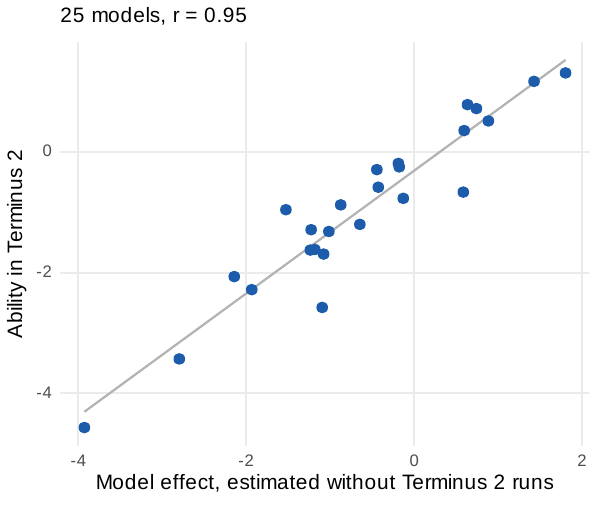}
\caption{Terminal-Bench 2.0: Model effects estimated without any Terminus 2 run, against the ability of the
same model's Terminus 2 run.}
\label{fig:terminus}
\end{figure}

\paragraph{A Fixed Reference Scaffold.} Because the same models were run in Terminus 2 and in other scaffolds,
the model effects can be checked against an external standard. We refitted the model without any Terminus 2
run and compared each model's effect with the ability of its Terminus 2 run, estimated by the Rasch model on
all runs. For the 25 models with both, the correlation was .95 (Spearman .93; Figure~\ref{fig:terminus}). A
model's effect learned from other scaffolds is thus close to its ability in a scaffold that held everything
else fixed, as \citet{ge2026agent} also found with variational estimates.

\paragraph{Predicting Held-Out Runs.} We left out each run in turn, refitted the model on the other 142 and
predicted the run from its model and scaffold. For the 102 runs whose model and scaffold had both been seen in
other runs, the predicted resolve rate was off by 4.1 percentage points on average, against 4.9 for the mean of
runs with the same model, 12.1 for the additive rule on rates and 18.4 for the grand mean, and the 95\%
intervals covered the run's ability 93\% of the time. With only one of the two seen (31 runs), the error was
10.9 points (coverage .90), and with neither (8 runs) it was 19.6 points.

\section{Discussion}

The model--scaffold question has a measurement answer and a design answer. The measurement answer is that
a cross-classified explanatory IRT model, fitted with standard mixed-model software in seconds to minutes,
separates the parts of an agent's ability as far as the data allow, tests additivity directly, and attaches
uncertainty to every number. The design answer is that the split is only trustworthy where models and
scaffolds cross. On all SWE-bench Verified runs, scaffolds seem to carry 38\% of the variance, but most of
those runs are the only evidence for both their model and their scaffold. Where the design crosses, both
leaderboards say the same thing: The language model accounts for 81\% to 91\% of the ability variance,
scaffolds for 5\% to 14\%, and the specific pair for 4\% to 5\%. Both reject strict additivity, in line with the
interaction contrasts of \citet{liu2026converged}, but the departure is small enough that the additive
decomposition of \citet{ge2026agent} predicts new pairs well, within 4 to 7 percentage points when both the
model and the scaffold have been seen.

\paragraph{What the Pair Term Means.} $\delta_a$ mixes three things: Genuine interaction (a scaffold tuned
to one model family), differences hidden by the labels (scaffold versions, attempt budgets, prompts), and
run-to-run noise. Repeated SWE-bench submissions with the same labels differed more than measurement error
allows, so the third and second kinds are part of it. A leaderboard that recorded the model, scaffold version
and attempt budget in a fixed form, as \citet{liu2026converged} also ask, and that kept attempt-level
results, would let the model separate these.

\paragraph{Data Quality.} Two of the 41 SWE-bench runs that report a score in their metadata had task-level
files that contradict it; the other 127 runs report none, so they could not be checked this way. Left in, they inflated the pair variance more than fivefold (from 0.29 to 1.67 logits$^2$), because
a run that scores zero is an extreme pair effect. Anyone fitting such models to public leaderboard files should
check task-level totals against reported scores first, and leaderboards should store the reported score
with every run.

\paragraph{Practical Advice.} Leaderboard maintainers who want to say what scaffolds contribute need
designs in which several scaffolds run the same models, ideally with repeated runs. A reference scaffold
such as mini-swe-agent, run with every new model, gives model comparisons that hold the scaffold fixed;
running a few strong models in every new scaffold would do the same for scaffolds. On both leaderboards, the choice of model matters far more than the choice of scaffold, but users should
still test a model in the scaffold they will deploy, because the pair term is never zero.

\paragraph{Limitations.} The data are observational and self-submitted: Teams choose models and scaffolds
together, so none of the effects is causal. The Rasch form fits worse than the 2PL, although the rankings
barely change; a cross-classified 2PL would let discrimination vary but is harder to fit. Tasks are scored
by the benchmark's tests, and public tasks may have leaked into training data \citep{liang2026illusion}. The two benchmarks differ in more
than their protocols (tasks, time span, models), so the contrast between them is suggestive rather than
a controlled comparison. The adaptive-testing check reuses runs that also set the difficulties, which flatters it
slightly; the held-out folds address this for prediction.

\section*{Software and Data Availability}

The analyses use R \citep{rcore2026}, \pkg{eirm} \citep{bulut2021eirm}, \pkg{lme4}
\citep{bates2015lme4}, \pkg{mirt} \citep{chalmers2012mirt} and \pkg{mirtverse}
\citep{bulut2026mirtverse}, available at \url{https://github.com/okanbulut/mirtverse}. The scripts that
reproduce every number and figure are in the repository's \code{paper-agents} folder. The SWE-bench results
are available from \url{https://github.com/SWE-bench/experiments}, and the Terminal-Bench 2.0 results from
\url{https://github.com/dariakryvosheieva/agent-psychometrics} (MIT license).

\section*{Use of Generative AI}

The author used Claude, an AI assistant from Anthropic, to help write the analysis code. The author designed the study, checked every analysis and number, and takes full responsibility for the content.

\bibliography{refs}

\end{document}